\documentclass{article}

\usepackage{microtype}
\usepackage{graphicx}
\usepackage{subfigure}
\usepackage{booktabs} 
\usepackage{algorithm}
\usepackage{algorithmic}
\usepackage{amsmath}

\usepackage[ruled,noresetcount,algo2e]{algorithm2e}

\usepackage{hyperref}

\usepackage[accepted]{icml2020}

\icmltitlerunning{Enhancing Neural Mathematical Reasoning by Abductive Combination with Symbolic Library}

\begin{document}

\twocolumn[
\icmltitle{Enhancing Neural Mathematical Reasoning by Abductive Combination with Symbolic Library}



\icmlsetsymbol{equal}{*}

\begin{icmlauthorlist}
\icmlauthor{Yangyang Hu}{equal,to}
\icmlauthor{Yang Yu}{equal,to}
\end{icmlauthorlist}

\icmlaffiliation{to}{State Key Laboratory for Novel Software Technology, Nanjing University, China. Polixir Technologies, China.}

\icmlcorrespondingauthor{Yangyang Hu}{yangyang.hu@polixir.ai}
\icmlcorrespondingauthor{Yang Yu}{yuy@nju.edu.cn}

\icmlkeywords{Machine Learning, ICML}

\vskip 0.3in
]



\printAffiliationsAndNotice{\icmlEqualContribution} 

\begin{abstract}
Mathematical reasoning recently has been shown as a hard challenge for neural systems. Abilities including expression translation, logical reasoning, and mathematics knowledge acquiring appear to be essential to overcome the challenge. This paper demonstrates that some abilities can be achieved through abductive combination with discrete systems that have been programmed with human knowledge. On a mathematical reasoning dataset, we adopt the recently proposed abductive learning framework, and propose the ABL-Sym algorithm that combines the Transformer neural models with a symbolic mathematics library. ABL-Sym shows 9.73\% accuracy improvement on the interpolation tasks and 47.22\% accuracy improvement on the extrapolation tasks, over the state-of-the-art approaches. Online demonstration: http://math.polixir.ai
\end{abstract}

\section{Introduction}


Automatically solving natural language described mathematical problems has been shown very challenging, requiring natural language understanding, mathematical expression extraction, and complex symbolic reasoning. Existing deep learning-related methods mainly frame these problems as a machine translation task. A branch of the methods explicitly encode the structural relation and try to directly output the answers~\citep{saxton2019analysing, schlag2019enhancing}. These methods have a great expression ability, but are hard to generalize to unseen cases. Another branch learns a mapping from the problem description to a solution program~\citep{wang2017deep,amini2019mathqa}, which explicitly encodes domain knowledge. These program-based methods rely heavily on human labeling, which is not only laborious, time-consuming, and error-prone. Besides, some problems are hard to be expressed in a program format, such as the varieties of probability problems (e.g., \texttt{Three letters picked without replacement from idiidauauuiuaiduaiiu. What is prob of sequence iaa?}).

Recently, the abductive learning~\citep{NIPS2019_8548} introduces a discrete logic module into a neural network with an integrated learning procedure. The logic module utilizes the logical consistency between the perception outputs and the logic background knowledge to optimize the perception module and the logic module jointly. This work demonstrates the possibility to produce a system with both the flexible perception power from neural networks and the generalization power from the programmed knowledge.     

In this paper, we follow the abductive learning framework and propose a system that integrates the transformer networks and a mathematical symbolic library, ABL-Sym, for automatically solving math problems. ABL-Sym firstly runs a consistency check and correction procedure: it generates programs from natural language descriptions and uses a program executor to run the programs; if the program output is inconsistent with the answer, it employs a search routine to correct the program. ABL-Sym then learns from the problem descriptions and the corrected programs. ABL-Sym repeats the two steps to improve its model. 
We evaluate ABL-Sym on the mathematics dataset from \citep{saxton2019analysing}. The results show that ABL-Sym significantly outperforms the previous state-of-the-art approaches: it achieves 9.72\% accuracy improvement on interpolation tasks, and 47.22\% accuracy improvement on extrapolation tasks.

\section{Background}
\subsection{The Mathematics Dataset}
Saxton et al. \citep{saxton2019analysing} introduced a mathematics dataset that contains a variety of math problems, including algebra, arithmetic, numerical comparison, numerical factorization, calculus, measurement, and probability. Each problem is a question-answer pair, where the question is like \texttt{Let q(m) = m**3 + 2. Let r(c) = -4*c**3 - 9. What is 18*q(f) + 4*r(f)?} and the answer is like \texttt{2*f**3}. Although there may be many forms of answer sequences with the same mathematical meaning, the evaluation criterion is character-by-character (i.e., each question is scored by either 0 or 1 according to whether the answer matches the correct answer character-by-character).  
The dataset is procedurally generated and consists of 56 modules, and each module provides 2M per-generated training samples and 10k interpolation samples. Extrapolation samples are also provided for an additional measure of algebraic generalization. 
  
\subsection{Sympy}
Sympy~\citep{meurer2017sympy} is a mathematical symbolic computing library, which contains about 300+ mathematical functions. Although many mathematical engines can be used, we adopt Sympy because it can conveniently get all the appropriate mathematical functions, easily exclude non-mathematical functions, and support direct access to the docstrings of mathematical functions. 

\subsection{Related work}
 
A mathematics dataset was released in \citep{saxton2019analysing} that analyzes the reasoning and generalization ability of popular reasoning neural architectures such as recurrent neural architectures and attention-augmented architectures (i.e., Transformer~\citep{vaswani2017attention}). The results show that the learned models did not do mathematical reasoning well, particularly for the extrapolation zone.
\citep{schlag2019enhancing} incorporates the tensor-product representation technique within the Transformer to better support the explicit representation of relation structure. They achieved improved results than the vanilla Transformer architecture without introducing any domain knowledge. 

Program format is a typical way to represent both of the discrete domain knowledge and the solution structure of mathematical problems. Amini et al. \citep{amini2019mathqa} released a dataset of math word problems that are densely annotated with programs by crowd-sourcing. Based on the dataset, they proposed a sequence-to-program model with automatic problem categorization. Comparing with their method, our approach applying to the dataset without annotated programs, and moreover, we use both the neural network and the discrete symbolic system for prediction. 


Abductive learning~\citep{NIPS2019_8548} was recently proposed for connecting a perception module with an abductive logical reasoning module using consistency optimization. The perception module generates output, the reasoning module checks and corrects the logical consistency, and the consistency information is used to update the perception module to generate logically more consistent output. This constitutes a forward cycle. Our approach is inspired by the above abductive learning framework, while we are addressing a different domain. 
 

\section{ABL-Sym}
In the following subsections, we introduce the program definition, the program correction, and the training procedure.

\subsection{Programs}

We define the program based on a domain-specific language (DSL) instead of arbitrary Turing-complete languages to reduce the search space of programs. Every word in the DSL is called an operator. All available operators form an operator space. The relationship between adjacent operators is appropriately restricted, such as \textit{argc} operators must be followed by \textit{math} operators, the number of optional variables must be no less than \textit{argc}, and \textit{argc} must be an available number of arguments to the mathematical operator.

\subsubsection{Operator Space}
The operator space consists of about 400 operators, including mathematical operators, position-aware operators, and several auxiliary operators.

\textbf {Mathematical Operators:}
We use Sympy as the program executor. In Sympy, there are about 300+ functions which are essential for solving math problems(e.g. \texttt{add}, \texttt{multiply}, \texttt{solve}, \texttt{diff}). We consider these functions as our mathematical operators.

\textbf{Position Operators: }
Mathematical expressions in problem may appear anywhere. We tokenize the problem sentence with a simple tokenizer and use positional indexes to identify expressions. The tokenizer uses space to tokenize the sentence and uses tokens that are not in the ordinary word dictionary as expressions. The ordinary word dictionary consists of non-digit words and excludes common ordinal number words (e.g., first, second, square). In addition, we also exclude a-z single letters because they are often used to represent variables in math problems. After tokenizing the problem, the positional continuous expression tokens are merged into one token. We use \texttt{pos0}, \texttt{pos1}, \texttt{pos2}, ... as positional operators to represent the positions of related expressions.

\textbf{Auxiliary Operators}
Functions in Sympy may have multiple parameters (e.g. \texttt{diff} function for obtaining derivative may have two usages: \texttt{diff(x**2+x\\*y, x)}, \texttt{diff(x**2+x*y, x, 2)}. We add \texttt{argc0}, \texttt{argc1}, \texttt{argc2}, \texttt{argc3} to the operator space in order to explicitly specify the number of function parameters. Some expressions in question do not conform to the input format of mathematical operators, and the output formats of some operators do not conform to the answer, so we add several additional format conversion operators and operator wrappers into the operator space.  

\subsubsection{Program Executor}
We build a simple program executor based on Sympy to run programs. In a running, the program's operators are executed sequentially, and intermediate results are saved in the environment through registry variables, which may be used by later operators. If an error is encountered during execution, execution will stop and return \texttt{none}, or if execution reaches the end, return execution $result$. 

\subsubsection{Programs Search procedure}
The program search space is too large to find the correct programs by random search. We design an abductive learning framework to search programs efficiently. Our framework performs multiple iterative searches. In the first iteration, we use a search-based method as a program generator to generate some programs. Then, the program executor runs the programs, and a consistency checker filters out the programs whose results are inconsistent with the answers. A neural network model is used to learn the mapping from the problem to the correct program. The learned model is then used to be a better program generator to start another iteration. 
In addition, we develop the following techniques to speed up the search process further. 

\paragraph{Warm-up Operator Distribution}
In math problems, problems are often strongly related to mathematical terms (e.g. in the derivative problems, the terms \texttt{derivative}, \texttt{differentiate} often appear). Additionally, almost every mathematical function in Sympy has a docstring, which usually contains related mathematical terms. So we can build relationships between problems and mathematical operators. In this paper, we adopt~\citep{arora2016simple} method to calculate the cosine similarity between the problem description and the docstring of an operator and then normalize by softmax to obtain the probability distribution of operators, which is used to generate the possible programs. 

\paragraph{Curriculum Search Strategy}
 According to whether the problem consists of simple problems, the problems in the Mathematics Dataset can be divided into simple problems and compositional problems (e.g. a compositional problem: \texttt{ Suppose -2*v + 1873 = 4*x - 3*x, x = 2*v - 1863. Let u = -65 + 25. Find the common denominator of 1/6 and v/(-920) - 8/u.}). Programs for simple problem can be found relatively easily by searching, but not for compositional problem. We observe that the compositional problem can be broken down into multiple parts, each of which is similar to a simple problem (e.g., the above problem can be broken down into three parts: \texttt{Suppose -2*v + 1873 = 4*x - 3*x, x = 2*v - 1863\#Let u = -65 + 25\#Find the common de\\nominator of 1/6 and v/(-920) - 8/u}). Therefore, we use the neural network model learned from simple problems to generate possible programs for each part and organize them into complete programs. The program executor then executes the programs to get results, and the consistency checker then checks the results for correctness. \\
 The whole search process is time-consuming, so we only perform search on randomly generated 500k problems that meet the qualifying conditions, and use the learned model to generate the rest.
   
\subsection{Neural Models}
The neural network model we use is a modified version of the original Transformer~\citep{vaswani2017attention}, with a shared transformer encoder $\theta^{enc}$ and two separate transformer decoders $\theta_{a}^{dec}$ and $\theta_{p}^{dec}$. We use the encoder $\theta^{enc}$ with hidden states $\mathbf{h}^{enc}$ to encode the problem $\mathbf{x}$. The decoders $\theta_{a}^{dec}$ and $\theta_{p}^{dec}$ take the shared hidden states $\mathbf{h}^{enc}$ and auto-regressively generates the answer sequence and program sequence respectively. During training, the decoders receive the shifted targets while during inference, we use the previously generated symbols with the highest probability. We treat the question and answer as a sequence of characters just like~\citep{vaswani2017attention} and treat the question as a sequence of operators. The overall training loss is the weighted sum of the answer decoding loss and the program decoding loss:
\begin{align*}
& \mathcal{L}(\theta^{enc},\theta_{a}^{dec}, \theta_{p}^{dec}) = \\
& -\alpha_{1}log P(\mathbf{y}_{a}|\mathbf{x};\theta^{enc},\theta_{a}^{dec}) -\alpha_{2}log
P(\mathbf{y}_{p}|\mathbf{x};\theta^{enc},\theta_{p}^{dec})
\end{align*}  

\begin{table*}[t!]
	\caption{Model accuracy averaged over all modules. A sample is correct if all characters of the target sequence have been predicted correctly. The column ``\textgreater 95\%'' counts how many of the 56 modules achieve over 95\% accuracy.}
	\label{tab:accurcy}
	\centering
	\setlength{\tabcolsep}{0.18cm}
	\begin{tabular}{lccccccc}
	\hline
		\multicolumn{1}{c}{} 
		& \multicolumn{1}{c}{weights} 
		& \multicolumn{1}{c}{steps}  
		& \multicolumn{2}{c}{interpolation} 
		& \multicolumn{2}{c}{extrapolation}
		\\
		\multicolumn{1}{c}{} 
		& 
		& 
		& \multicolumn{1}{c}{acc} & \multicolumn{1}{c}{\textgreater 95\%}
		& \multicolumn{1}{c}{acc} & \multicolumn{1}{c}{\textgreater 95\%}
		\\
		\hline
		Transformer (Saxton et al.)  & 30M      & 500k  & 76.00\% & 13 & 50.00\% & 1 \\
		TP-Transformer (Schlag et al.)              & 49.2M    & 700k  & 80.67\% & 18 & 52.48\% & 3 \\
		\hline
		Transformer (ours)           & 44.2M    & 700k  & 76.41\% & 13 & 50.48\% & 2 \\
		TP-Transformer (ours)        & 49.2M    & 700k  & 79.82\% & 18 & 51.99\% & 3 \\
		\hline
		ABL-Sym+Transformer (ours)               & 54.9M    & 700k  & \bf 87.85\% & \bf 29 & \bf 73.41\% & \bf 7 \\
		ABL-Sym+TP-Transformer (ours)            & 58.8M    & 700k  & \bf 88.52\% & \bf 33 & \bf 77.26\% & \bf 8 \\\hline
	\end{tabular}
\end{table*}

\section{Experiments}
We evaluate our framework on the mathematics dataset \citep{saxton2019analysing}. The reason we did not evaluate on other mathematical datasets~\citep{kushman2014learning,huang2016well,upadhyay2016annotating,wang2017deep,ling2017program,amini2019mathqa} is because these datasets are either limited to narrow specific fields or demanded for manual annotated programs. 

\subsection{Settings}
During the search, the maximum number of sampled programs for each problem is $N_{w}=100k$ on the first iteration and $N_{n}=1k$ on the other iterations. The number of iterations $I$ is set to 5.

We extract a character-level vocabulary of 72 symbols and an operator-level vocabulary of 380 symbols, both including \texttt{START}, \texttt{END}, and \texttt{PADDING} symbols. 

Our transformer-like model parameters $\theta^{enc}$, $\theta_{a}^{dec}$, $\theta_{p}^{dec}$ are set to an embedding size of $512$, with $8$ attentional heads, and intermediate feed-forward dimension of 2048. The answer decoder $\theta_{a}^{dec}$ is with layers of $6$ while the program decoder $\theta_{p}^{dec}$ is with layers of $2$. We train our model via the Adam optimizer ~\citep{kingma2014adam} with a learning rate of $8 \times 10^{-5}$, $\beta_1 = 0.9$, $\beta_2 = 0.995$, $\epsilon = 10^{-9}$. We use a batch size of $1024$, with absolute gradient value clipping of $0.1$. We trained our model on one server with 8 V100 Nvidia GPUs for 12 days.
During the search process, the parameters configuration of our program-generated model is the same as the above model.

At the inference, answers and programs are generated by sequential decoding. If the predicted program is \texttt{none} or fails to run successfully, the neural model answer is used as the final result.  

\subsection{Experimental Results}

\begin{figure}[t!]
	\centering
	\includegraphics[width=\linewidth]{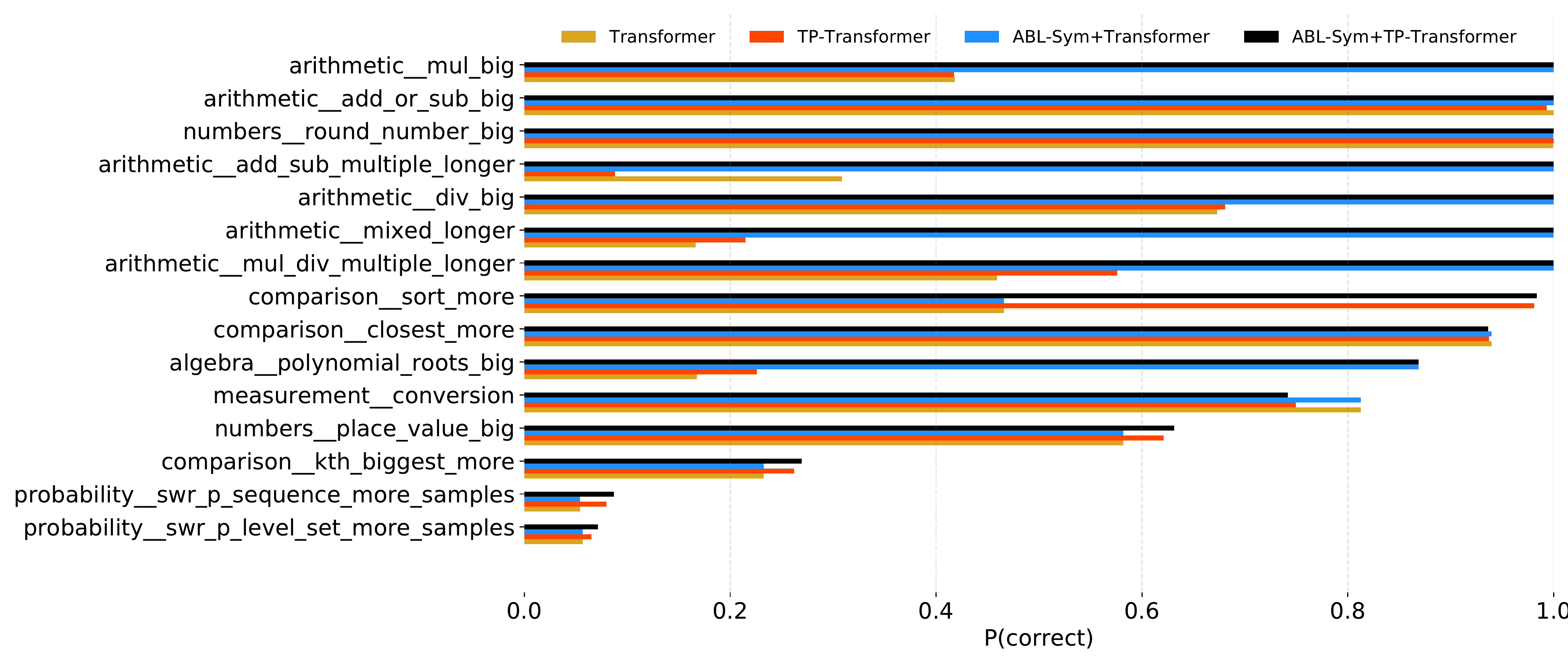}
	\caption{The extrapolation test performances of our implementations of Transformer, TP-Transformer(700K steps) and our ABL-Sym framework based on Transformer and TP-Transformer on the different modules.}
	\label{fig:performance_extrapolate}
\end{figure}

Table~\ref{tab:accurcy} presents the overall performance on the dataset. We can see that our model significantly outperforms the previous state-of-the-art by up to  $7.8\%$ absolute improvement on the interpolation test dataset and $24.8\%$ absolute improvement on the extrapolation test dataset. Our program-augmented model dramatically improves the performance of the model, especially for generalizing the model to areas not previously seen. For a more detailed comparison, Fig.~\ref{fig:performance_extrapolate} shows the test performance on extrapolation modules.

Table~\ref{tab:search} shows the performance of the 5 iterations of ABL-Sym together with the random search performance. ABL-Sym shows clearly better than random search. In the first iteration, it used an average of 20\% fewer search times than the random search strategy but found 76\% more programs, which mainly due to the warm-up strategy and curriculum search strategy. These strategies allow us to search more programs faster within the maximum search limit. After the first iteration, the model we learned as a better program generator generated better candidate programs, so we searched an additional 8\% of the programs with negligible search times. Compared to the second iteration, the number of programs searched in the next few iterations increased by only a litter bit. This is because most programs that can be searched are also almost searched. Still, $57.7\%$ programs were not found.

\begin{table}[t!]
	\caption{The cost and the hit ratio of programs during iterating}
	\label{tab:search}
	\centering
	\setlength{\tabcolsep}{0.18cm}
	\begin{tabular}{lccccccc}
	\hline
		\multicolumn{1}{c}{Method} 
		& \multicolumn{1}{c}{per-question searches} 
		& \multicolumn{1}{c}{hit ratio}  
		\\
		\hline
  		ABL-Sym (1 itr)              & 64.14k    & 33.2\%  \\
		ABL-Sym (2 itrs)              & 64.86k    &  40.1\%  \\
		ABL-Sym (3 itrs)              & 65.49k    &  41.3\%  \\
		ABL-Sym (4 itrs)              & 66.11k    &  42.0\%  \\
		ABL-Sym (5 itrs)              & 66.73k    & \bf 42.3\%  \\
				\hline
		Random search            & 82.09k       & 18.9\%       \\	
		\hline
	\end{tabular}
\end{table}

ABL-Sym can find many programs of compositional or complex problems (e.g., \texttt{Calculate the common denominator of 25/13728 and 121/1248.} the program found is \texttt{pos7 argc1 denom pos5 argc1 denom argc2 lcm}), but random search strategy was failed.

\section{Conclusion}
In this work, we demonstrate that integrating discrete systems into neural systems is a feasible way to enhance the neural systems, particularly in the extrapolation ability.  Notice that even human beings learn complex knowledge, e.g. mathematics, progressively from well organized textbooks. Well designed discrete systems may serve the role of textbooks for building a complex intelligent systems.

\bibliographystyle{icml2020}
\bibliography{dmmath}


\end{document}